\pdfoutput=1
\documentclass[11pt]{article}

\usepackage[preprint]{acl}

\usepackage{times}
\usepackage{latexsym}
\usepackage[T1]{fontenc}
\usepackage[utf8]{inputenc}
\usepackage{microtype}
\usepackage{inconsolata}

\usepackage{graphicx}
\usepackage{booktabs}
\usepackage{tabularx}
\usepackage{colortbl}
\usepackage{multirow}
\usepackage{longtable}
\usepackage{array}

\usepackage{enumerate}
\usepackage{pdflscape}
\usepackage{float}
\usepackage[most]{tcolorbox}

\usepackage{hyperref}
\usepackage{url}

\title{Classification is a RAG problem: A case study on hate speech detection}

\author{
  \textbf{Richard Willats}$^{1}$ \quad
  \textbf{Josh Pennington}$^{1}$ \quad
  \textbf{Aravind Mohan}$^{1}$ \thanks{Email: aravind.mohan@contextual.ai} \quad
  \textbf{Bertie Vidgen}$^{1}$ \\
  $^1$Contextual AI 
}

\begin{document}
\maketitle

\begin{abstract}
Robust content moderation requires classification systems that can quickly adapt to evolving policies without costly retraining. We present classification using Retrieval-Augmented Generation (RAG), which shifts traditional classification tasks from determining the correct category in accordance with pre-trained parameters to evaluating content in relation to contextual knowledge retrieved at inference. In hate speech detection, this transforms the task from \textit{``is this hate speech?''} to \textit{``does this violate the hate speech policy?''}

Our \textbf{\textsc{Contextual Policy Engine (CPE)}} -- an agentic RAG system -- demonstrates this approach and offers three key advantages: (1) robust classification accuracy comparable to leading commercial systems, (2) inherent explainability via retrieved policy segments, and (3) dynamic policy updates without model retraining. Through three experiments, we demonstrate strong baseline performance and show that the system can apply fine-grained policy control by correctly adjusting protection for specific identity groups without requiring retraining or compromising overall performance. These findings establish that RAG can transform classification into a more flexible, transparent, and adaptable process for content moderation and wider classification problems.
\end{abstract}

\section{Introduction}

\begin{figure*}[t]
    \centering
    \includegraphics[width=\textwidth]{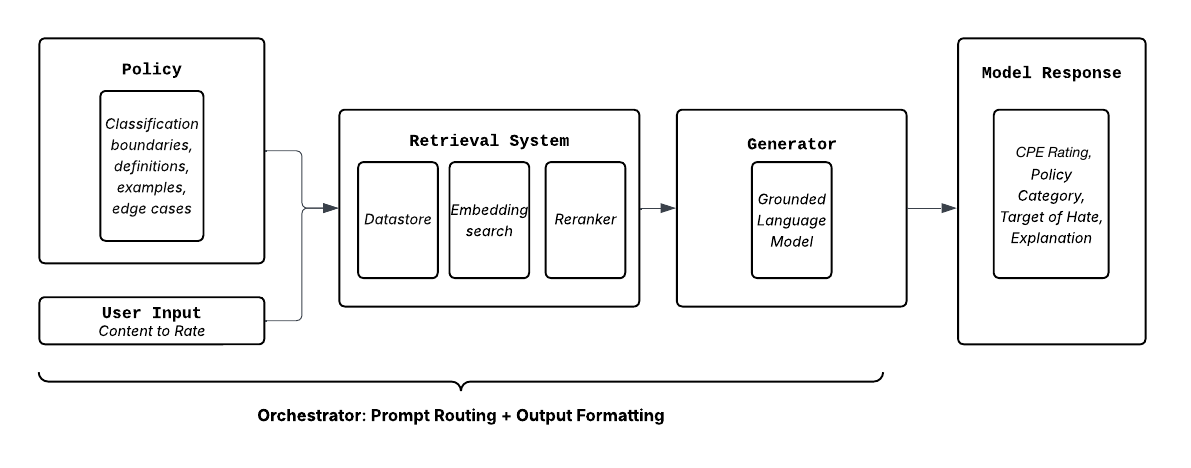}
    \caption{Architecture of the Contextual Policy Engine (CPE).}
    \label{fig:oracle-diagram}
\end{figure*}

Supervised machine learning classifiers automatically categorize data into predefined categories. From predicting customer churn to assessing review sentiment and moderating social content, there are numerous business, consumer, and social applications. 

We present an alternative approach to classification using Retrieval-Augmented Generation (RAG). RAG is widely used to improve performance at question-answering: a RAG system reads the user input, retrieves relevant information from a knowledge store (usually identified through semantic and keyword matching), and then provides this to the generator so it has more context to reason over. We use this approach to improve classification by giving a generative model more context in the form of relevant policy documentation. This helps solve key limitations of generative AI classifiers, such as hallucinations, inconsistency, and brittleness. It also offers greater flexibility, as the system can be updated by refreshing the documents rather than by parametric retraining. 

The key benefits of using RAG for classification include:

\begin{enumerate}
   \item \textbf{Improved performance.} RAG provides a principled way of making classifications, providing the model with access to the exact information it needs to assess the content correctly. This can produce higher-quality results and is, in principle, more generalizable to unseen content. 
   \item \textbf{Inherently explainable.} The retrieved evidence used by the system can be exposed to the user, providing a precise \textit{ante-hoc} explanation for classifications. Given that these retrievals can be long and hard to read, they can be summarized by another model to return a free-text explanation. 
   \item \textbf{Easy to steer and update.} A RAG system requires zero training or tuning to deliver SOTA results (although fine-tuning can deliver additional performance benefits on top). If the documents are updated, the system can immediately use them to update how it classifies. This enables customized classification to meet the needs of specific users, as well as various axes (such as territories, teams, and segments) without any retraining. 
\end{enumerate}

We introduce the \textbf{\textsc{Contextual Policy Engine (CPE)}}, a new system built on top of State-of-the-Art components from Contextual AI, without any training. The CPE provides document retrievals, policy categories, and explanations with each classification, augmenting human moderation work. It is available in beta at \url{https://huggingface.co/spaces/rwillats/guardrails}.

We use the CPE for a case study on hate classification, a challenging problem with applications in content moderation and trust \& safety. We address the problem of \textit{adjustable} hate speech detection, whereby a hate speech policy is updated to adjust which identity groups receive protection based on evolving standards. This is a live problem for many social media companies, whether they regularly update their policies or not, and drives the need for systems that can implement changes without costly retraining. We run three experiments and demonstrate that the CPE is extensible and adjustable. It can enforce policy changes that extend protection to additional identity groups, or remove protection from others, with minimal reduction in performance.

\section{RAG classification system}
\label{sec:ragsystem}

The CPE classifies content the way a human would---by reading the content, finding relevant policies, and determining the appropriate label. This matches better with industry-norms as it transforms classification from assessing the correct category for the content \textit{in general} to evaluating the content \textit{in relation to specific documents}. In our hate speech case study, the task shifts from ``Is this hate speech?'' to ``Does this violate the hate speech policy?''

The RAG classification system has four key components:

\begin{enumerate}
\item \textbf{Policy.} Describes the criteria for the classification categories (for the case study: Within Policy or Out of Policy). The Policy must be detailed, explicit, and comprehensive, containing definitions, explanations, exemplars, and edge cases. We encourage a prescriptive approach to Policy creation and refinement where the category boundaries are precisely defined. Otherwise, limitations in the Policy will result in misclassifications.

\item \textbf{Retrieval system.} Retrieves relevant content from the Policy for the content that is being assessed. The system includes a datastore with chunked documents, embedding search, and reranking to select appropriate chunks. We use Contextual AI's SOTA retrieval and reranking system.\footnote{See: \url{https://contextual.ai/blog/introducing-instruction-following-reranker/}.}

\item \textbf{Generator.} This component processes the content being assessed and the retrieved content to generate a response. Our implementation uses Contextual AI's Grounded Language Model, which adheres to document information rather than parametric knowledge and provides strong reasoning capabilities. It is preference-optimized from Llama 3.3.\footnote{See: \url{https://contextual.ai/blog/introducing-grounded-language-model/}.}

\item \textbf{Orchestrator.} Combines system prompts with user input and retrieved knowledge. It instructs the generator to return classifications, policy categories, and explanations based on retrievals. The orchestrator can be calibrated to optimize for recall, precision, or specific types of input.
\end{enumerate}

\subsection{Contextual Policy Engine}

The CPE was deployed with access to a detailed hate speech policy authored by in-house safety experts. The system was prompted with zero-shot instructions to classify content as either Within Policy or Out of Policy. The policy defines explicit classification boundaries for protected identities and types of hate, including dehumanization, discrimination, and incitement to violence or harm. These definitions are accompanied by explanations and edge-case examples to support consistent interpretation. By grounding judgments in retrieved policy content, the CPE enables evidence-based classification with transparent reasoning, as illustrated in Figure~\ref{fig:oracle-diagram}.

\section{Experiment 1: Performance of Systems Under Test (SUTs) at classifying content}

We evaluate SUT performance in classifying user-generated social content as Hateful or Non-Hateful.

\subsection{Systems Under Test}

We compare the CPE against three widely-used content moderation systems: (1) LlamaGuard 3 (8B), (2) OpenAI's content moderation API, and (3) The Perspective API from Google Jigsaw. All systems were accessed between April and May 2025 using their default configurations.

To ensure fair comparison, we evaluated each SUT under two conditions: (1) All harm categories and (2) only hate-specific categories: \textit{``S10: Hate Speech''} for LlamaGuard, \textit{``hate''} and \textit{``hate/threatening''} for OpenAI, and \textit{``identity attack''} for Perspective API. We used a cut-off of 0.5 on the confidence scores to determine whether content is Hateful or Non-Hateful. In total, we evaluate each of the three commercial systems under two conditions, plus the CPE, resulting in seven SUTs (additional system configuration details provided in Appendix~\ref{app:sut_methodology}).

\subsection{Labeled evaluation dataset}

\begin{table}[htp]
\centering
\small
\renewcommand{\arraystretch}{1.2}
\setlength{\tabcolsep}{8pt}
\begin{tabular}{lrr}
\toprule
\textbf{Target Identity} & \textbf{Non-Hateful} & \textbf{Hateful} \\
\midrule
Black people & 125 & 357 \\
Disabled people & 111 & 373 \\
Gay people & 178 & 373 \\
Immigrants & 106 & 357 \\
Muslims & 111 & 373 \\
Trans people & 106 & 357 \\
Women & 136 & 373 \\
No target & 292 & 0 \\
\midrule
\textbf{Total} & \textbf{1,165} & \textbf{2,563} \\
\bottomrule
\end{tabular}
\caption{Distribution of HateCheck dataset by target identity (n=3,728)}
\label{tab:hatecheck_distribution}
\end{table}

We evaluate the SUTs against the HateCheck dataset \citep{rottger-etal-2021-hatecheck}, a granular test suite designed to evaluate hate speech detection models across various functional categories of hate and non-hate such as reclaimed slurs, negated hate, counter speech, derogation, and dehumanization. We selected HateCheck based on the following: (1) it provides clean, well-defined labels with minimal noise, (2) it covers seven distinct protected identity groups with consistent labeling, (3) its template-based test cases can be adapted for new identities, (4) it distinguishes between different functional types of hate speech, and (5) each test case has secondary labels indicating targeted groups and whether the hate is directed at groups or individuals.

\subsection{Experiment 1 Results}

\begin{table}[htbp]
\centering
\scriptsize
\setlength{\tabcolsep}{0pt}
\renewcommand{\arraystretch}{1.2}
\begin{tabular*}{\columnwidth}{@{\extracolsep{\fill}}l@{\hspace{1.5pt}}c@{\hspace{1pt}}c@{\hspace{1pt}}c@{\hspace{1pt}}c@{\hspace{1pt}}c@{\hspace{1pt}}c@{\hspace{1pt}}c@{\hspace{1pt}}c@{}}
\toprule
\textbf{Model} & \textbf{F1} & \textbf{Acc} & \textbf{Prec} & \textbf{Rec} & \textbf{TP} & \textbf{FP} & \textbf{TN} & \textbf{FN} \\
\midrule
\rowcolor{gray!20} Contextual Policy Engine & \textbf{0.988} & 0.984 & 0.983 & 0.993 & 2529 & 43 & 1122 & 171 \\
\midrule
OpenAI-Default & 0.925 & 0.889 & 0.861 & 1.0 & 2563 & 413 & 752 & 0 \\
\rowcolor{gray!20} OpenAI-Hate & \textbf{0.996} & 0.994 & 0.991 & 1.0 & 2563 & 23 & 1142 & 0 \\
\midrule
LlamaG-Default & 0.936 & 0.916 & 0.974 & 0.901 & 2310 & 61 & 1104 & 253 \\
LlamaG-Hate & 0.887 & 0.859 & 0.985 & 0.807 & 2069 & 31 & 1134 & 494 \\
\midrule
Persp-Default & 0.864 & 0.802 & 0.818 & 0.915 & 2345 & 521 & 644 & 218 \\
Persp-Hate & 0.851 & 0.804 & 0.893 & 0.812 & 2081 & 249 & 916 & 482 \\
\bottomrule
\end{tabular*}
\caption{Performance comparison on the HateCheck dataset (n=3,728). Systems with F1 scores above 0.95 are highlighted.}
\label{tab:hatecheck_model_comparison}
\end{table}

Both \textbf{OpenAI-Hate} and the \textbf{Contextual Policy Engine} achieve the highest performance, with F1 scores of \textbf{0.996} and \textbf{0.988} respectively. The worst performing SUT is Perspective-Hate, with an F1 score of 0.851. As anticipated, the commercial systems that are limited to only hate-specific categories have higher precision than all-category conditions. However, for two of the SUTs, recall is lower as well. Despite this, all SUTs show strong performance, indicating that this is a tractable task on standard benchmarks.

While these results demonstrate strong baseline capabilities, benchmark performance alone may not reflect real-world deployment scenarios where content distribution and policy requirements can vary significantly. The adaptability of classification systems to specific domains and evolving policies represents an important consideration beyond initial performance metrics.

The SUTs also differ in their output formats and the information provided. For the commercial SUTs, ratings were returned as binary flags or scaled scores (0--1) for overall assessment and specific harm category violations. The CPE returns a structured response that includes the classification label, the type and target of hate, and a free-text explanation grounded in retrieved policy content. An example output from Experiment 1 is shown in Figure~\ref{fig:oracle-example}.

\begin{figure}[ht]
\centering
\begin{tcolorbox}[colback=gray!5!white, colframe=gray!75!black]
\textbf{Content to Rate:} Muslims belong in a zoo.

\vspace{0.5em}
\textbf{Content Rating:} Out of Policy

\textbf{Policy Category:} Dehumanization

\textbf{Target of Hate:} Religion (Muslims)

\textbf{Explanation:} This content directly compares Muslims to animals, which is a clear form of dehumanization. The policy explicitly defines such comparisons as denying human dignity and reinforcing prejudice against protected identities. Muslims are a protected group under the religion category, making this comparison a direct policy violation.
\end{tcolorbox}
\caption{Example output from the \textsc{Contextual Policy Engine (CPE)}, demonstrating its transparent, policy-grounded classification.}
\label{fig:oracle-example}
\end{figure}

\section{Experiment 2: Adding new targets of hate}

\subsection{Experimental Setup}

We added three new targets of hate to HateCheck. These targets are not always included in definitions of hate speech and have typically received less attention in existing approaches. 

\begin{itemize}
    \item \textbf{Trump voters.} Voting status falls outside traditional protected characteristics in most hate speech frameworks. However, it represents an important dimension of identity, and political belief is increasingly being recognized for protection in some contexts.
    \item \textbf{Furries.} Members of this subculture identify with anthropomorphic animal characters, sometimes in personal or sexual ways. They are frequently subjected to targeted online harassment, including the use of specific derogatory terms, and the community itself documents numerous slurs used against them \citep{wikifur2023}.
    \item \textbf{Homeless people.} Socioeconomic status is rarely explicitly protected in hate speech policies despite documented patterns of dehumanizing language targeting homeless individuals on social media \citep{pardo2020violence}.
\end{itemize}

HateCheck was created with templates, which allows us to slot in new targets of hate programmatically. We created 460 test cases for each group (354 hateful, 106 non-hateful) for a total of 1,380 test cases. Each new case was reviewed by one of the study authors, and no issues were identified. The terms used in the templates are given in Appendix~\ref{app:added_targets}. For each identity, we simply added these identity groups to the protected identities section of the Policy for the CPE. We made no changes to the commercial SUTs.

\begin{table}[!htbp]
\centering
\small
\begin{tabular}{lrrr}
\toprule
\textbf{Dataset} & \textbf{Total N} & \textbf{Non-hateful} & \textbf{Hateful} \\
\midrule
\textbf{Total} & \textbf{1,380} & \textbf{318} & \textbf{1,062} \\
Trump voters & 460 & 106 & 354 \\
Furries & 460 & 106 & 354 \\
Homeless people & 460 & 106 & 354 \\
\bottomrule
\end{tabular}
\caption{Distribution of the extended identity test sets}
\label{tab:extended_identity_distribution}
\end{table}

\subsection{Experiment 2 Results}

As shown in Table~\ref{tab:extended_identity_results}, the CPE achieves the highest F1 (0.972) across the combined test sets for the extended identity groups. Compared with the SUTs' performance in Experiment 1, the CPE records the lowest drop in F1 score of only 1.6\% (from 0.988 to 0.972). In contrast, commercial SUTs show significantly larger performance drops: OpenAI's HateSpeech-only configuration drops 7.3\% (from 0.996 to 0.923), LlamaGuard-Hate drops 52.6\% (from 0.887 to 0.420), and Perspective-Hate experiences the most severe degradation at 83.2\% (from 0.851 to 0.143).

OpenAI automatically extends some protection to all three new identity groups, though with reduced effectiveness. LlamaGuard and Perspective generally do not recognize these non-traditional groups as protected (with the partial exception of LlamaGuard's rating of the attacks on homeless people), resulting in dramatically reduced performance. With all three commercial solutions, users have limited control over these protection boundaries -- they do not have access to the model parameters or policies and, even if they do, extending or removing protection would require costly and time-consuming model retraining, and possibly new data labeling. 

\begin{table}[ht]
\centering
\scriptsize
\begin{tabular*}{\columnwidth}{@{\extracolsep{\fill}}lr@{\hspace{12pt}}r@{\hspace{6pt}}@{}}
\toprule
\textbf{Model} & \textbf{F1} & \textbf{Acc.} \\
\midrule
\multicolumn{3}{l}{\cellcolor{gray!15}\textbf{HateCheck added identities (total) (n=1380)}} \\
\midrule
Contextual Policy Engine - Hate Speech & \textbf{0.972} & \textbf{0.957} \\
Open AI Moderation - Default config & 0.968 & 0.949 \\
Open AI Moderation - HateSpeech & 0.923 & 0.890 \\
LlamaGuard-8b - Default config & 0.597 & 0.554 \\
LlamaGuard-8b - HateSpeech & 0.420 & 0.433 \\
Perspective - Default config & 0.771 & 0.694 \\
Perspective - HateSpeech & 0.143 & 0.290 \\
\midrule
\multicolumn{3}{l}{\cellcolor{gray!15}\textbf{Trump voters (n=460)}} \\
\midrule
Contextual Policy Engine - Hate Speech & 0.947 & 0.920 \\
Open AI Moderation - Default config & \textbf{0.970} & \textbf{0.952} \\
Open AI Moderation - HateSpeech & 0.894 & 0.852 \\
LlamaGuard-8b - Default config & 0.464 & 0.457 \\
LlamaGuard-8b - HateSpeech & 0.211 & 0.317 \\
Perspective - Default config & 0.799 & 0.724 \\
Perspective - HateSpeech & 0.086 & 0.265 \\
\midrule
\multicolumn{3}{l}{\cellcolor{gray!15}\textbf{Furries (n=460)}} \\
\midrule
Contextual Policy Engine - Hate Speech & \textbf{0.979} & \textbf{0.967} \\
Open AI Moderation - Default config & 0.963 & 0.941 \\
Open AI Moderation - HateSpeech & 0.873 & 0.826 \\
LlamaGuard-8b - Default config & 0.497 & 0.480 \\
LlamaGuard-8b - HateSpeech & 0.329 & 0.378 \\
Perspective - Default config & 0.746 & 0.670 \\
Perspective - HateSpeech & 0.208 & 0.320 \\
\midrule
\multicolumn{3}{l}{\cellcolor{gray!15}\textbf{Homeless people (n=460)}} \\
\midrule
Contextual Policy Engine - Hate Speech & 0.990 & 0.985 \\
Open AI Moderation - Default config & 0.971 & 0.954 \\
Open AI Moderation - HateSpeech & \textbf{0.994} & \textbf{0.991} \\
LlamaGuard-8b - Default config & 0.784 & 0.724 \\
LlamaGuard-8b - HateSpeech & 0.651 & 0.602 \\
Perspective - Default config & 0.767 & 0.689 \\
Perspective - HateSpeech & 0.132 & 0.285 \\
\bottomrule
\end{tabular*}
\caption{Performance on extended identity test sets}
\label{tab:extended_identity_results}
\end{table}

\section{Experiment 3: Adjustable hate speech detection}

\subsection{Experimental Setup}

To assess how well the CPE handles policy adjustments for different protected identities, we created three new variant evaluation sets for the three identities used in Experiment 2 (Trump voters, Furries, and Homeless people). To do this, we: (1) selected one identity to exempt from protection, (2) kept the other two identities as protected, (3) relabeled all 354 previously hateful cases targeting the exempted identity as ``Non-Hateful'', and (4) maintained the original 106 non-hateful cases for each identity. This process resulted in datasets containing 672 non-hateful cases (354 exempted cases + 318 original non-hateful cases) and 708 hateful cases (354 cases for each of the two protected identities).

For each variant, we modified the CPE's policy by removing the selected identity from the list of protected groups. OpenAI Moderation, LlamaGuard, and Perspective could not be evaluated in this experiment, as their APIs do not allow for customization of protected categories without access to their underlying models or system-level integration.

\subsection{Experiment 3 Results}

\begin{table*}[t]
\centering
\small
\renewcommand{\arraystretch}{1.2}
\setlength{\tabcolsep}{3.5pt}
\begin{tabular}{cl|cc|cc|cc}
\toprule
\multirow{2}{*}{\textbf{\#}} & \multirow{2}{*}{\textbf{Dataset}} & \multicolumn{2}{c|}{\textbf{Trump voters}} & \multicolumn{2}{c|}{\textbf{Furries}} & \multicolumn{2}{c}{\textbf{Homeless people}} \\
\cmidrule{3-8}
 & & \textbf{Non-hate (n/\%)} & \textbf{Hate (n/\%)} & \textbf{Non-hate (n/\%)} & \textbf{Hate (n/\%)} & \textbf{Non-hate (n/\%)} & \textbf{Hate (n/\%)} \\
\midrule
\rowcolor{gray!15}
1 & Original & 90/106 & 333/354 & 96/106 & 349/354 & 103/106 & 350/354 \\
\rowcolor{gray!15}
 & & (84.91\%) & (94.07\%) & (90.57\%) & (98.59\%) & (97.17\%) & (98.87\%) \\
\midrule
2 & Trump voters & \textbf{447/460} & \textbf{0/0} & 95/106 & 346/354 & 102/106 & 351/354 \\
 & exempt & \textbf{(97.17\%)} & \textbf{(-)} & (89.62\%) & (97.74\%) & (96.23\%) & (99.15\%) \\
\midrule
3 & Furries & 80/106 & 306/354 & \textbf{456/460} & \textbf{0/0} & 103/106 & 349/354 \\
 & exempt & (75.47\%) & (86.44\%) & \textbf{(99.13\%)} & \textbf{(-)} & (97.17\%) & (98.59\%) \\
\midrule
4 & Homeless people & 90/106 & 301/354 & 98/106 & 347/354 & \textbf{453/460} & \textbf{0/0} \\
 & exempt & (84.91\%) & (85.03\%) & (92.45\%) & (98.02\%) & \textbf{(98.48\%)} & \textbf{(-)} \\
\bottomrule
\end{tabular}
\caption{Detection accuracy when exempting specific identity groups from protection. Bold cells indicate exempted groups.}
\label{tab:identity_exemption_results}
\end{table*}

Table~\ref{tab:identity_exemption_results} demonstrates the CPE's ability to selectively apply classification boundaries based on protected status applied to identities in the policy document. When we exempt specific identities from the policy, the CPE performs well at adhering to the new instructions and ensuring the correct classification boundary is maintained. We find that (1) attacks against the identity that has been excluded from protection are correctly classified as Non-hateful; and (2) classification of attacks against the other identities are mostly maintained, though with some notable degradation. While many cases show minimal impact (<2\%), we observed more significant drops in some configurations, particularly with Trump voters where performance decreased by approximately 10\%.

The results show different patterns across the three identity groups. When \textbf{Trump voters} are exempted from protection, the model achieves a true negative rate of 97.17\% for previously hateful content targeting this group. This exemption has minimal impact on false negative rates for other protected groups, with hateful content detection remaining robust for furries (97.74\%) and homeless people (99.15\%). When \textbf{Furries} are exempted, the model shows the highest true negative accuracy (99.13\%) among all exemption scenarios. However, this comes with a notable increase in false negatives for Trump voter content (86.44\% detection rate compared to the original 94.07\%). Similarly, when \textbf{Homeless people} are exempted, the model maintains high true negative performance (98.48\%), but significantly increases false negatives for Trump voter content (85.03\% detection rate). This increase in false negatives likely stems from overlapping definitions or terminology in the policy documents rather than inherent linguistic patterns.

\section{Conclusion}

The CPE demonstrates that RAG presents an effective approach for machine learning classification. It offers greater performance, explainability, and consistency. Our hate speech case study demonstrates that the CPE achieves competitive performance at a difficult classification task. We also demonstrate that the system is flexible and adjustable, with zero training. Importantly, this solution can be used for any expert human knowledge work. It presents a powerful way of augmenting and supporting the work of subject matter experts. 

While our approach demonstrates significant advantages for policy-driven classification, several limitations should be acknowledged:

\begin{itemize}
    \item \textbf{Policy.} Because RAG systems require access to a set of documents, our approach exposes any gaps in the documentation - if the documents are not complete or poorly written, the system cannot give a correct classification regardless of model capabilities. This challenge can be addressed by iteratively reviewing classifications, comparing against the documents, and plugging any gaps.
    \item \textbf{Retrieval.} When policy documents are extensive, retrieval quality becomes a potential bottleneck. Complex queries may not retrieve the most relevant policy sections, affecting classification accuracy.
    \item \textbf{Computational cost.} The RAG approach introduces additional computational costs compared to pure parametric classification, with retrieval and reasoning steps that may impact latency in high-throughput applications.
\end{itemize}

This work has introduced several interesting avenues for future research, such as training a system for RAG-based classification, further experimentation with more targets of hate and evalsets, and evaluation of the explanations. Feedback on the demo is welcome. To use the Contextual Policy Engine in production, reach out to the study authors.

\section{Previous work}

Hate speech detection, classification and monitoring has been extensively researched for both user-generated social content and user interactions with AI models. While many labelled datasets have been introduced to train and evaluate models, numerous challenges have been identified in hate speech detection, such as: (1) the role of context in determining whether content is hateful, such as the social setting, conversation, and person speaking \citep{vidgen-etal-2021-introducing, markov-daelemans-2022-role, fleisig-etal-2023-majority}; (2) the subjective nature of assessing hate, whereby different individuals construe the same content differently \citep{rottger-etal-2022-two, das2024investigatingannotatorbiaslarge}; (3) the difficulty of assessing content in non-English languages and non-text modalities \citep{ousidhoum-etal-2019-multilingual, mathias-etal-2021-findings, rottger-etal-2022-multilingual, haber-etal-2023-improving}; and (4) the lexical, syntactic and semantic complexity of real-world hate \citep{schmidt-wiegand-2017-survey, 10.1371/journal.pone.0243300}. These factors make it difficult for hate detection systems to perform well and be trusted when used in production.

Extensive work has also focused on explainability in hate speech detection, which typically involves providing fine-grained classification, such as detecting specific targets and types of hate, as well as providing free-text rationales for classifications. For instance, \citet{kirk-etal-2023-semeval} introduce the explainable sexism detection task at SemEval 2023. They present a three-tiered hierarchical labelling framework, with the third tier offering a classification for one of 11 distinct sexism vectors. Similarly, \citet{mathias-etal-2021-findings} relabel a dataset of hateful memes for the vector and target of hate. \citet{elsherief-etal-2021-latent} introduce a benchmark that provides free-text explanations of the `implication' of hateful statements, as well as finegrained secondary labels. \citet{yang-etal-2023-hare} use an LM to improve the annotation schemas used by annotators (and LMs) to label hate. This helps to improve how models perform at identifying hate, and quality of their auto-generated free-text rationales.

While RAG has been primarily used to improve LM performance at question answering and natural language reasoning \citep{lewis2021retrievalaugmentedgenerationknowledgeintensivenlp, shuster-etal-2021-retrieval-augmentation, es2023ragasautomatedevaluationretrieval, fan2024surveyragmeetingllms, gao2024retrievalaugmentedgenerationlargelanguage}, increasingly, RAG has been combined with agentic approaches that enable agents to be stateful and take actions based on external data sources, APIs, and other inputs \citep{lála2023paperqaretrievalaugmentedgenerativeagent, song2023llmplannerfewshotgroundedplanning, skarlinski2024languageagentsachievesuperhuman, singh2025agenticretrievalaugmentedgenerationsurvey}. A few early studies have explored using RAG for classification tasks. Class-RAG, introduced by Meta's GenAI team \citep{chen2024classragrealtimecontentmoderation}, applies RAG to content moderation by retrieving relevant examples to guide classification decisions. They demonstrate adaptability to external datasets and instruction following through experiments with modified retrieval libraries, showing their system can flip classifications when safety labels are reversed in the knowledge base. Building on these promising results, our work extends this concept by implementing a complete RAG-based classification framework that enables more targeted and fine-grained policy modifications without sacrificing overall system performance.

\bibliography{custom}

\begin{thebibliography}{30}
\providecommand{\natexlab}[1]{#1}

\bibitem[{Chen et~al.(2024)Chen, Shen, Bavalatti, Lin, Wang, Hu, Subramanyam, Vepuri, Jiang, Qi, Chen, Jiang, and Jain}]{chen2024classragrealtimecontentmoderation}
Jianfa Chen, Emily Shen, Trupti Bavalatti, Xiaowen Lin, Yongkai Wang, Shuming Hu, Harihar Subramanyam, Ksheeraj~Sai Vepuri, Ming Jiang, Ji~Qi, Li~Chen, Nan Jiang, and Ankit Jain. 2024.
\newblock \href {https://arxiv.org/abs/2410.14881} {Class-rag: Real-time content moderation with retrieval augmented generation}.
\newblock \emph{Preprint}, arXiv:2410.14881.

\bibitem[{Das et~al.(2024)Das, Zhang, Hasan, Sarkar, Jamshidi, Bhattacharya, Rahgouy, Raychawdhary, Feng, Jain, Chadha, Sandage, Pope, Dozier, and Seals}]{das2024investigatingannotatorbiaslarge}
Amit Das, Zheng Zhang, Najib Hasan, Souvika Sarkar, Fatemeh Jamshidi, Tathagata Bhattacharya, Mostafa Rahgouy, Nilanjana Raychawdhary, Dongji Feng, Vinija Jain, Aman Chadha, Mary Sandage, Lauramarie Pope, Gerry Dozier, and Cheryl Seals. 2024.
\newblock \href {https://arxiv.org/abs/2406.11109} {Investigating annotator bias in large language models for hate speech detection}.
\newblock \emph{Preprint}, arXiv:2406.11109.

\bibitem[{ElSherief et~al.(2021)ElSherief, Ziems, Muchlinski, Anupindi, Seybolt, De~Choudhury, and Yang}]{elsherief-etal-2021-latent}
Mai ElSherief, Caleb Ziems, David Muchlinski, Vaishnavi Anupindi, Jordyn Seybolt, Munmun De~Choudhury, and Diyi Yang. 2021.
\newblock \href {https://doi.org/10.18653/v1/2021.emnlp-main.29} {Latent hatred: A benchmark for understanding implicit hate speech}.
\newblock In \emph{Proceedings of the 2021 Conference on Empirical Methods in Natural Language Processing}, pages 345--363, Online and Punta Cana, Dominican Republic. Association for Computational Linguistics.

\bibitem[{Es et~al.(2023)Es, James, Espinosa-Anke, and Schockaert}]{es2023ragasautomatedevaluationretrieval}
Shahul Es, Jithin James, Luis Espinosa-Anke, and Steven Schockaert. 2023.
\newblock \href {https://arxiv.org/abs/2309.15217} {Ragas: Automated evaluation of retrieval augmented generation}.
\newblock \emph{Preprint}, arXiv:2309.15217.

\bibitem[{Fan et~al.(2024)Fan, Ding, Ning, Wang, Li, Yin, Chua, and Li}]{fan2024surveyragmeetingllms}
Wenqi Fan, Yujuan Ding, Liangbo Ning, Shijie Wang, Hengyun Li, Dawei Yin, Tat-Seng Chua, and Qing Li. 2024.
\newblock \href {https://arxiv.org/abs/2405.06211} {A survey on rag meeting llms: Towards retrieval-augmented large language models}.
\newblock \emph{Preprint}, arXiv:2405.06211.

\bibitem[{Fleisig et~al.(2023)Fleisig, Abebe, and Klein}]{fleisig-etal-2023-majority}
Eve Fleisig, Rediet Abebe, and Dan Klein. 2023.
\newblock \href {https://doi.org/10.18653/v1/2023.emnlp-main.415} {When the majority is wrong: Modeling annotator disagreement for subjective tasks}.
\newblock In \emph{Proceedings of the 2023 Conference on Empirical Methods in Natural Language Processing}, pages 6715--6726, Singapore. Association for Computational Linguistics.

\bibitem[{Gao et~al.(2024)Gao, Xiong, Gao, Jia, Pan, Bi, Dai, Sun, Wang, and Wang}]{gao2024retrievalaugmentedgenerationlargelanguage}
Yunfan Gao, Yun Xiong, Xinyu Gao, Kangxiang Jia, Jinliu Pan, Yuxi Bi, Yi~Dai, Jiawei Sun, Meng Wang, and Haofen Wang. 2024.
\newblock \href {https://arxiv.org/abs/2312.10997} {Retrieval-augmented generation for large language models: A survey}.
\newblock \emph{Preprint}, arXiv:2312.10997.

\bibitem[{Haber et~al.(2023)Haber, Vidgen, Chapman, Agarwal, Lee, Yap, and R{\"o}ttger}]{haber-etal-2023-improving}
Janosch Haber, Bertie Vidgen, Matthew Chapman, Vibhor Agarwal, Roy Ka-Wei Lee, Yong~Keong Yap, and Paul R{\"o}ttger. 2023.
\newblock \href {https://doi.org/10.18653/v1/2023.acl-long.711} {Improving the detection of multilingual online attacks with rich social media data from {S}ingapore}.
\newblock In \emph{Proceedings of the 61st Annual Meeting of the Association for Computational Linguistics (Volume 1: Long Papers)}, pages 12705--12721, Toronto, Canada. Association for Computational Linguistics.

\bibitem[{{Jigsaw}(2024)}]{perspectiveapi}
{Jigsaw}. 2024.
\newblock About the api: Attributes and languages.
\newblock \url{https://support.perspectiveapi.com/s/about-the-api-attributes-and-languages}.
\newblock Accessed: April 2025.

\bibitem[{Kirk et~al.(2023)Kirk, Yin, Vidgen, and R{\"o}ttger}]{kirk-etal-2023-semeval}
Hannah Kirk, Wenjie Yin, Bertie Vidgen, and Paul R{\"o}ttger. 2023.
\newblock \href {https://doi.org/10.18653/v1/2023.semeval-1.305} {{S}em{E}val-2023 task 10: Explainable detection of online sexism}.
\newblock In \emph{Proceedings of the 17th International Workshop on Semantic Evaluation (SemEval-2023)}, pages 2193--2210, Toronto, Canada. Association for Computational Linguistics.

\bibitem[{Lewis et~al.(2021)Lewis, Perez, Piktus, Petroni, Karpukhin, Goyal, Küttler, Lewis, tau Yih, Rocktäschel, Riedel, and Kiela}]{lewis2021retrievalaugmentedgenerationknowledgeintensivenlp}
Patrick Lewis, Ethan Perez, Aleksandra Piktus, Fabio Petroni, Vladimir Karpukhin, Naman Goyal, Heinrich Küttler, Mike Lewis, Wen tau Yih, Tim Rocktäschel, Sebastian Riedel, and Douwe Kiela. 2021.
\newblock \href {https://arxiv.org/abs/2005.11401} {Retrieval-augmented generation for knowledge-intensive nlp tasks}.
\newblock \emph{Preprint}, arXiv:2005.11401.

\bibitem[{Lála et~al.(2023)Lála, O'Donoghue, Shtedritski, Cox, Rodriques, and White}]{lála2023paperqaretrievalaugmentedgenerativeagent}
Jakub Lála, Odhran O'Donoghue, Aleksandar Shtedritski, Sam Cox, Samuel~G. Rodriques, and Andrew~D. White. 2023.
\newblock \href {https://arxiv.org/abs/2312.07559} {Paperqa: Retrieval-augmented generative agent for scientific research}.
\newblock \emph{Preprint}, arXiv:2312.07559.

\bibitem[{Markov and Daelemans(2022)}]{markov-daelemans-2022-role}
Ilia Markov and Walter Daelemans. 2022.
\newblock \href {https://aclanthology.org/2022.trac-1.5/} {The role of context in detecting the target of hate speech}.
\newblock In \emph{Proceedings of the Third Workshop on Threat, Aggression and Cyberbullying (TRAC 2022)}, pages 37--42, Gyeongju, Republic of Korea. Association for Computational Linguistics.

\bibitem[{Mathias et~al.(2021)Mathias, Nie, Mostafazadeh~Davani, Kiela, Prabhakaran, Vidgen, and Waseem}]{mathias-etal-2021-findings}
Lambert Mathias, Shaoliang Nie, Aida Mostafazadeh~Davani, Douwe Kiela, Vinodkumar Prabhakaran, Bertie Vidgen, and Zeerak Waseem. 2021.
\newblock \href {https://doi.org/10.18653/v1/2021.woah-1.21} {Findings of the {WOAH} 5 shared task on fine grained hateful memes detection}.
\newblock In \emph{Proceedings of the 5th Workshop on Online Abuse and Harms (WOAH 2021)}, pages 201--206, Online. Association for Computational Linguistics.

\bibitem[{{Meta AI}(2024)}]{meta_llama_guard}
{Meta AI}. 2024.
\newblock Llama guard 3 model card and prompt format.
\newblock \url{https://www.llama.com/docs/model-cards-and-prompt-formats/llama-guard-3/}.
\newblock Accessed: April 2025.

\bibitem[{{OpenAI}(2024)}]{openai_moderation}
{OpenAI}. 2024.
\newblock Moderation.
\newblock \url{https://platform.openai.com/docs/guides/moderation}.
\newblock Accessed: April 2025.

\bibitem[{Ousidhoum et~al.(2019)Ousidhoum, Lin, Zhang, Song, and Yeung}]{ousidhoum-etal-2019-multilingual}
Nedjma Ousidhoum, Zizheng Lin, Hongming Zhang, Yangqiu Song, and Dit-Yan Yeung. 2019.
\newblock \href {https://doi.org/10.18653/v1/D19-1474} {Multilingual and multi-aspect hate speech analysis}.
\newblock In \emph{Proceedings of the 2019 Conference on Empirical Methods in Natural Language Processing and the 9th International Joint Conference on Natural Language Processing (EMNLP-IJCNLP)}, pages 4675--4684, Hong Kong, China. Association for Computational Linguistics.

\bibitem[{Pardo(2020)}]{pardo2020violence}
Maria~Laura Pardo. 2020.
\newblock Violence and hate speech against the homeless in social media during the covid-19 pandemic.
\newblock In Alexandra Cotoc, Octavian More, and Mihaela Mudure, editors, \emph{Multicultural Discourses in Turbulent Times}, pages 191--210. Presa Universitară Clujeana.

\bibitem[{R{\"o}ttger et~al.(2022{\natexlab{a}})R{\"o}ttger, Seelawi, Nozza, Talat, and Vidgen}]{rottger-etal-2022-multilingual}
Paul R{\"o}ttger, Haitham Seelawi, Debora Nozza, Zeerak Talat, and Bertie Vidgen. 2022{\natexlab{a}}.
\newblock \href {https://doi.org/10.18653/v1/2022.woah-1.15} {Multilingual {H}ate{C}heck: Functional tests for multilingual hate speech detection models}.
\newblock In \emph{Proceedings of the Sixth Workshop on Online Abuse and Harms (WOAH)}, pages 154--169, Seattle, Washington (Hybrid). Association for Computational Linguistics.

\bibitem[{R{\"o}ttger et~al.(2022{\natexlab{b}})R{\"o}ttger, Vidgen, Hovy, and Pierrehumbert}]{rottger-etal-2022-two}
Paul R{\"o}ttger, Bertie Vidgen, Dirk Hovy, and Janet Pierrehumbert. 2022{\natexlab{b}}.
\newblock \href {https://doi.org/10.18653/v1/2022.naacl-main.13} {Two contrasting data annotation paradigms for subjective {NLP} tasks}.
\newblock In \emph{Proceedings of the 2022 Conference of the North American Chapter of the Association for Computational Linguistics: Human Language Technologies}, pages 175--190, Seattle, United States. Association for Computational Linguistics.

\bibitem[{R{\"o}ttger et~al.(2021)R{\"o}ttger, Vidgen, Nguyen, Waseem, Margetts, and Pierrehumbert}]{rottger-etal-2021-hatecheck}
Paul R{\"o}ttger, Bertie Vidgen, Dong Nguyen, Zeerak Waseem, Helen Margetts, and Janet Pierrehumbert. 2021.
\newblock \href {https://doi.org/10.18653/v1/2021.acl-long.4} {{H}ate{C}heck: Functional tests for hate speech detection models}.
\newblock In \emph{Proceedings of the 59th Annual Meeting of the Association for Computational Linguistics and the 11th International Joint Conference on Natural Language Processing (Volume 1: Long Papers)}, pages 41--58, Online. Association for Computational Linguistics.

\bibitem[{Schmidt and Wiegand(2017)}]{schmidt-wiegand-2017-survey}
Anna Schmidt and Michael Wiegand. 2017.
\newblock \href {https://doi.org/10.18653/v1/W17-1101} {A survey on hate speech detection using natural language processing}.
\newblock In \emph{Proceedings of the Fifth International Workshop on Natural Language Processing for Social Media}, pages 1--10, Valencia, Spain. Association for Computational Linguistics.

\bibitem[{Shuster et~al.(2021)Shuster, Poff, Chen, Kiela, and Weston}]{shuster-etal-2021-retrieval-augmentation}
Kurt Shuster, Spencer Poff, Moya Chen, Douwe Kiela, and Jason Weston. 2021.
\newblock \href {https://doi.org/10.18653/v1/2021.findings-emnlp.320} {Retrieval augmentation reduces hallucination in conversation}.
\newblock In \emph{Findings of the Association for Computational Linguistics: EMNLP 2021}, pages 3784--3803, Punta Cana, Dominican Republic. Association for Computational Linguistics.

\bibitem[{Singh et~al.(2025)Singh, Ehtesham, Kumar, and Khoei}]{singh2025agenticretrievalaugmentedgenerationsurvey}
Aditi Singh, Abul Ehtesham, Saket Kumar, and Tala~Talaei Khoei. 2025.
\newblock \href {https://arxiv.org/abs/2501.09136} {Agentic retrieval-augmented generation: A survey on agentic rag}.
\newblock \emph{Preprint}, arXiv:2501.09136.

\bibitem[{Skarlinski et~al.(2024)Skarlinski, Cox, Laurent, Braza, Hinks, Hammerling, Ponnapati, Rodriques, and White}]{skarlinski2024languageagentsachievesuperhuman}
Michael~D. Skarlinski, Sam Cox, Jon~M. Laurent, James~D. Braza, Michaela Hinks, Michael~J. Hammerling, Manvitha Ponnapati, Samuel~G. Rodriques, and Andrew~D. White. 2024.
\newblock \href {https://arxiv.org/abs/2409.13740} {Language agents achieve superhuman synthesis of scientific knowledge}.
\newblock \emph{Preprint}, arXiv:2409.13740.

\bibitem[{Song et~al.(2023)Song, Wu, Washington, Sadler, Chao, and Su}]{song2023llmplannerfewshotgroundedplanning}
Chan~Hee Song, Jiaman Wu, Clayton Washington, Brian~M. Sadler, Wei-Lun Chao, and Yu~Su. 2023.
\newblock \href {https://arxiv.org/abs/2212.04088} {Llm-planner: Few-shot grounded planning for embodied agents with large language models}.
\newblock \emph{Preprint}, arXiv:2212.04088.

\bibitem[{Vidgen and Derczynski(2021)}]{10.1371/journal.pone.0243300}
Bertie Vidgen and Leon Derczynski. 2021.
\newblock \href {https://doi.org/10.1371/journal.pone.0243300} {Directions in abusive language training data, a systematic review: Garbage in, garbage out}.
\newblock \emph{PLOS ONE}, 15(12):1--32.

\bibitem[{Vidgen et~al.(2021)Vidgen, Nguyen, Margetts, Rossini, and Tromble}]{vidgen-etal-2021-introducing}
Bertie Vidgen, Dong Nguyen, Helen Margetts, Patricia Rossini, and Rebekah Tromble. 2021.
\newblock \href {https://doi.org/10.18653/v1/2021.naacl-main.182} {Introducing {CAD}: the contextual abuse dataset}.
\newblock In \emph{Proceedings of the 2021 Conference of the North American Chapter of the Association for Computational Linguistics: Human Language Technologies}, pages 2289--2303, Online. Association for Computational Linguistics.

\bibitem[{{WikiFur}(2023)}]{wikifur2023}
{WikiFur}. 2023.
\newblock \href {https://en.wikifur.com/wiki/Category:Derogatory_terms} {Category:derogatory terms}.
\newblock Accessed: 2025-06-10.

\bibitem[{Yang et~al.(2023)Yang, Kim, Kim, Ho, Thorne, and Yun}]{yang-etal-2023-hare}
Yongjin Yang, Joonkee Kim, Yujin Kim, Namgyu Ho, James Thorne, and Se-Young Yun. 2023.
\newblock \href {https://doi.org/10.18653/v1/2023.findings-emnlp.365} {{HARE}: Explainable hate speech detection with step-by-step reasoning}.
\newblock In \emph{Findings of the Association for Computational Linguistics: EMNLP 2023}, pages 5490--5505, Singapore. Association for Computational Linguistics.

\end{thebibliography}

\appendix

\section{Appendix: Systems Under Test Methodology}
\label{app:sut_methodology}

\subsection*{Llama Guard 3}
We used Meta's Llama-Guard-3-8B checkpoint accessed via the Together API in April 2025. Our implementation followed Meta's official prompt format~\citep{meta_llama_guard}. The model was queried with default temperature settings. We provide results for the model calculated by binary safety judgments (``safe'' or ``unsafe'') and more fine-grained results based on the model labelling the input ``S10: Hate Speech''.

\subsection*{OpenAI Moderation}
We used OpenAI's Moderation API with the ``omni-moderation-latest'' model and default settings through OpenAI's official Python SDK~\citep{openai_moderation} in April 2025. We provide results for the model calculated by binary safety judgments (flagged: ``true'' or ``false'') and results based on the model flagging ``true'' for ``hate'' or ``hate/threatening'' classification.

\subsection*{Perspective API}
We used the Google Perspective API and requested scores for the five production attributes, and omitted the experimental attributes according to the model card~\citep{perspectiveapi}. We provide binary classifications of unsafe, calculated where any of the attributes scores are equal or greater than 0.5. We also provide results only for hate speech detection, calculated where the ``IDENTITY\_ATTACK'' attribute score is greater than 0.5.

\section{Appendix: Additional Targets of Hate}
\label{app:added_targets}

\begin{table}[htbp]
\centering
\begin{tabular}{|l|l|l|l|}
\hline
\rowcolor{gray!15}\textbf{Identity} & \textbf{Alternative Term} & \textbf{Slur 1} & \textbf{Slur 2} \\
\hline
Trump supporter & MAGA & Trumptard & MAGAT \\
\hline
Furry & Anthropomorphic & Furfag & Furvert \\
\hline
Homeless person & Unhoused person & Gutter trash & Street rat \\
\hline
\end{tabular}
\caption{Identity terms used in extended test cases for Experiment 2}
\label{tab:appendix_identity_terms}
\end{table}

\end{document}